\documentclass[10pt]{article} % For LaTeX2e

\usepackage[accepted]{rlj} % Should be uncommented for the camera-ready
\usepackage{amssymb}            % Defines common symbols like \mathbb R
\usepackage{mathtools}          % Extends amsmath, providing common math tools
\usepackage{mathrsfs}           % Enables \mathscr, which can work in cases that \mathcal does not
\usepackage{graphicx}           % For including images
\usepackage{subcaption}         % Allows for the use of subfigures and subcaptions
\usepackage[space]{grffile}     % For spaces in image names
\usepackage{url}                % For displaying URLs
\usepackage{lipsum}             % For placeholder text
\usepackage{booktabs}
\usepackage{caption}

\title{In-Distribution Imagination for Model-Based Offline Reinforcement Learning
}

\setrunningtitle{In-Distribution Imagination}

\author{Mintae Kim, Koushil Sreenath}

\emails{\{mintae.kim, koushils\}@berkeley.edu}

\affiliations{
\textbf{\textit{Hybrid Robotics}, BAIR, UC Berkeley}
}

\contribution{
    We propose IDI, a rollout control framework for MBORL that estimates trajectory support in a learned latent trajectory space and adaptively truncates rollouts that leave the offline trajectory manifold.

    }
    {
    Existing MBORL methods primarily control rollouts using transition-level uncertainty estimates or introduce pessimism to models or critics, whereas IDI performs trajectory-level rollout control using learned trajectory support.
    }

\contribution{
    We show that rollout reliability is better characterized by trajectory support than by transition-level uncertainty. In learned dynamics rollouts, trajectory support exhibits a substantially stronger relationship with rollout error than ensemble disagreement.
    }
    {
    Existing uncertainty-based rollout control methods assume that local transition uncertainty is a reliable indicator of long-horizon rollout quality.
    }

\contribution{
    We demonstrate that trajectory-level rollout control improves the quality of imagined data and consistently improves performance in limited-data MBORL.
    }
    {
    Experiments are conducted in a limited-data regime where synthetic trajectories are necessary to compensate for reduced offline coverage.
    }

 % Your keywords

\summary{Model-based offline reinforcement learning (MBORL) improves sample efficiency by generating synthetic trajectories from learned dynamics models. However, accumulative model error can gradually drive imagined trajectories outside the offline data distribution, producing unrealistic synthetic data and unstable policy optimization. Existing MBORL methods primarily control rollouts using transition-level uncertainty estimates, implicitly assuming that local prediction reliability reflects long-horizon rollout quality.

This paper proposes \emph{in-distribution imagination} (IDI), a trajectory-level rollout control framework that estimates trajectory support in a learned latent trajectory space and adaptively truncates rollouts that leave the offline trajectory manifold. IDI is combined with trajectory-regularized actor-critic (TRAC), a dataset-aware extension of entropy-regularized RL. We evaluate the proposed method in a limited-data setting where model-generated trajectories are necessary to compensate for reduced offline coverage.

Experimental results show that trajectory support predicts rollout failure substantially better than transition-level uncertainty and that IDI consistently improves offline RL performance across MuJoCo locomotion benchmarks. These results suggest that reliable imagination is fundamentally a trajectory-level problem and highlight the importance of trajectory-level rollout control for MBORL.
}

\begin{document}

% \makeCover  % Create the cover page
\maketitle  % Make the title section

\begin{abstract}
Model-based offline reinforcement learning (MBORL) improves sample efficiency through model-generated trajectories.
However, accumulative model error can drive imagined trajectories outside the offline data distribution, leading to unrealistic synthetic data and unstable policy optimization.
Many existing methods primarily control rollouts using transition-level uncertainty.
We propose \emph{in-distribution imagination} (IDI), a rollout control framework that estimates trajectory support in a learned representation space and adaptively truncates rollouts that leave the offline trajectory manifold.
Combined with trajectory-regularized RL, an extension of entropy-regularized RL, IDI consistently improves performance in limited-data settings.
Experiments show that trajectory support predicts rollout failure substantially better than transition-level uncertainty, highlighting the importance of trajectory-level rollout control in MBORL.
\end{abstract}

%%%%%%%%%%%%%%%%%%%%%%%%%%%%%%%%%%%%%%%%%%%%%%%%%%%%%%%%%%%%%%%%
%% Section: Introduction
%%%%%%%%%%%%%%%%%%%%%%%%%%%%%%%%%%%%%%%%%%%%%%%%%%%%%%%%%%%%%%%%

\section{Introduction}
\label{sec:introduction}

Offline reinforcement learning (RL) learns policies from static datasets without environment interaction.
Model-based offline RL (MBORL) further improves sample efficiency by learning dynamics models and generating synthetic trajectories for policy optimization.
However, accumulative model error often drives imagined trajectories outside the offline data distribution, allowing policies to exploit model inaccuracies and leading to unstable policy learning \cite{yu2020mopo, kim2026wombet}.
Existing MBORL methods primarily control rollouts through transition-level uncertainty estimation and pessimistic imagination.
Common approaches penalize rewards via step-wise uncertainty estimation, terminate rollouts in uncertain regions, or introduce pessimism to models or critics \cite{yu2020mopo, kidambi2020morel, yu2021combo, lyu2022double, zheng2023model, kim2026robust}.
While effective, these methods assess rollout quality using local transition reliability or prevent value overestimation by not fully trusting synthetic data for critic learning.
A key observation of this work is that rollout reliability is a property of trajectories rather than individual transitions: a sequence of locally plausible transitions may still accumulate into an unrealistic trajectory.
Thus, transition-level uncertainty alone may be insufficient to determine rollout reliability.

\begin{figure}[t]
    
    \centering
    \includegraphics[width=0.8\linewidth]{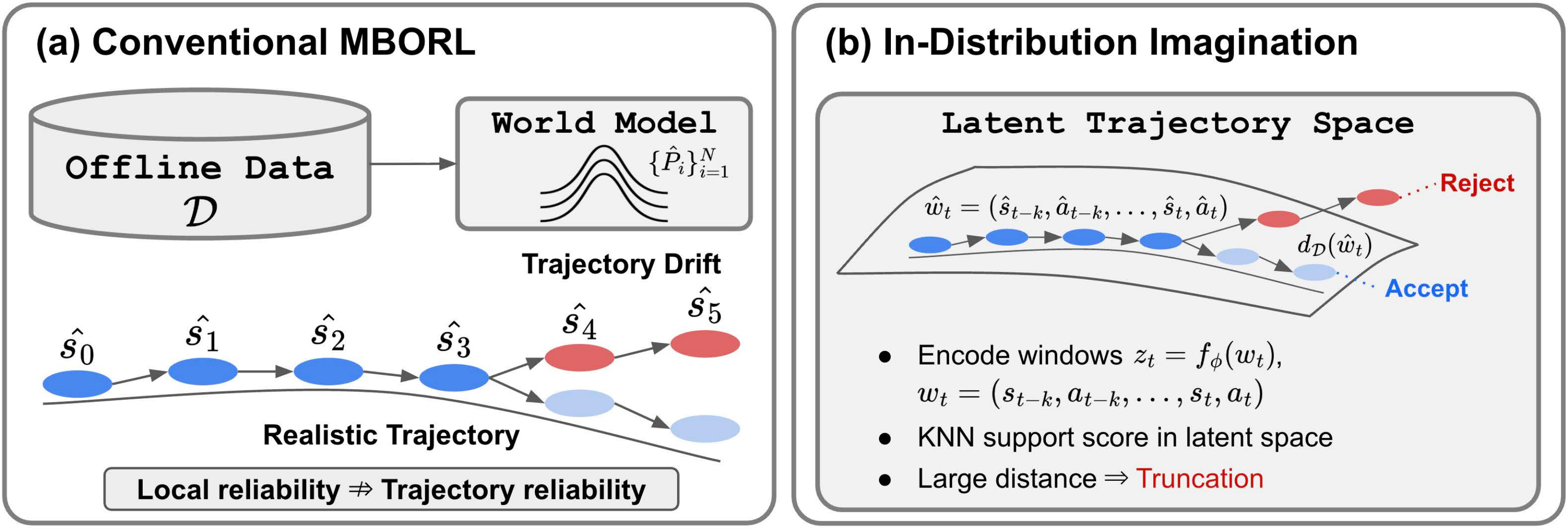}
    \caption{
    Conceptual overview of in-distribution imagination (IDI).
    (a) Conventional MBORL relies on transition-level reliability, but locally plausible transitions may still accumulate into unrealistic trajectories due to compounding model error.
    (b) IDI instead measures trajectory support in a learned latent trajectory space and truncates imagination when trajectories leave the offline manifold.
    This converts rollout control from local uncertainty estimation to trajectory-level support estimation.
    }
    \label{fig:concept}
    
\end{figure}

We propose \emph{in-distribution imagination} (IDI), a trajectory-level rollout control framework for MBORL.
IDI learns trajectory representations from offline data, estimates support using a $K$-nearest-neighbor (KNN) score in the learned latent space, and adaptively truncates rollouts that leave the offline trajectory manifold.
As a result, imagination continues only while generated trajectories remain supported by offline trajectories.
Figure~\ref{fig:concept} illustrates the key intuitions behind IDI.
We combine IDI with \emph{trajectory-regularized actor-critic} (TRAC), an offline RL framework that regularizes the policy-induced trajectory distribution toward the offline trajectory distribution.
The two components operate at complementary stages of learning: IDI improves imagined data quality, while TRAC improves policy optimization using the data from IDI and thereby mitigating distributional shift.
Since standard locomotion benchmarks can often be solved without synthetic data, we evaluate IDI in a limited-data regime where model-generated trajectories are genuinely useful.
Experiments show that trajectory support predicts rollout failure substantially better than transition-level uncertainty and that IDI consistently improves offline RL performance.
These results suggest that reliable imagination requires trajectory-level rollout control.

Our contributions are summarized as follows:
\begin{itemize}
\item We identify rollout reliability as a trajectory-level property.
\item We propose IDI, a trajectory-level rollout control framework based on latent-space support.
\item We show that trajectory support predicts rollout failure better than transition-level uncertainty and improves overall MBORL performance.
\end{itemize}

%%%%%%%%%%%%%%%%%%%%%%%%%%%%%%%%%%%%%%%%%%%%%%%%%%%%%%%%%%%%%%%%
%% Section: Preliminaries
%%%%%%%%%%%%%%%%%%%%%%%%%%%%%%%%%%%%%%%%%%%%%%%%%%%%%%%%%%%%%%%%

\section{Preliminaries}

We consider an infinite-horizon discounted Markov decision process (MDP) $(\mathcal S,\mathcal A,P,r,\gamma,\rho_0)$, where $\mathcal S$ is the state space, $\mathcal A$ is the action space, $P(s'|s,a)$ denotes the transition dynamics, $r(s,a)$ is the reward function, $\gamma \in [0,1)$ is the discount factor, and $\rho_0$ is the initial state distribution.
In offline RL, the agent learns solely from a static dataset $\mathcal D=\{\tau_i\}_{i=1}^N$, where each trajectory is $\tau=(s_0,a_0,r_0,s_1,a_1,r_1,\dots,s_T)$.
The objective is to learn a policy $\pi(a|s)$ maximizing the discounted return $J(\pi)=\mathbb E_{\pi, P}\left[\sum_{t=0}^{\infty}\gamma^t r(s_t,a_t)\right]$.
MBORL learns a transition dynamics model $\hat P_\theta(s'|s,a)$ from $\mathcal D$ and uses it to generate imagined trajectories for policy optimization.
Starting from states sampled from $\mathcal D$, imagined trajectories are generated by $\hat a_t\sim\pi(\cdot|\hat s_t)$ and $\hat s_{t+1}\sim\hat P_\theta(\cdot|\hat s_t,\hat a_t)$, yielding imagined trajectories $\hat\tau=(\hat s_0,\hat a_0,\hat s_1,\hat a_1,\dots)$.
Imagined transitions generated from imagined trajectories are then combined with offline data for policy optimization.
MBORL typically models transition dynamics using a probabilistic ensemble.
Specifically, an ensemble of $N$ dynamics models
$\{\hat P_i\}_{i=1}^N$
is trained on the offline dataset, where each model predicts a Gaussian next-state distribution
$\hat P_i(s'|s,a)=\mathcal N(\mu_i(s,a),\Sigma_i(s,a))$.
The ensemble provides both state predictions and uncertainty estimates.

%%%%%%%%%%%%%%%%%%%%%%%%%%%%%%%%%%%%%%%%%%%%%%%%%%%%%%%%%%%%%%%%
%% Section: In-Distribution Imagination for Trajectory-Regularized Offline RL
%%%%%%%%%%%%%%%%%%%%%%%%%%%%%%%%%%%%%%%%%%%%%%%%%%%%%%%%%%%%%%%%

\section{In-Distribution Imagination for Trajectory-Regularized Offline RL}
\label{sec:idi-trac}

We propose \emph{in-distribution imagination} (IDI), a history-aware rollout control framework for MBORL.
Rather than evaluating each imagined transition independently, IDI measures whether the transition remains compatible with the recent rollout history.

For each transition $\xi_t=(s_t,a_t,s_{t+1})$, we construct a window containing the most recent $k$ transitions:
\begin{equation}
w_t
=
(\xi_{t-k+1},\ldots,\xi_t)
=
(s_{t-k+1},a_{t-k+1},\ldots,s_t,a_t,s_{t+1}).
\label{eq:trajectory_window}
\end{equation}
The inclusion of $s_{t+1}$ ensures that the support score evaluates the newly generated transition before it is accepted.
We encode each window as $z_t=f_\phi(w_t)$.
The encoder is trained using only offline trajectory windows and remains fixed during model rollout generation.

The encoder is optimized via contrastive learning \cite{oord2018representation, qiao2025sumo, wang20261000}.
Given an anchor window $w_i$, positives are sampled from nearby windows within the same trajectory, while negatives are sampled from unrelated trajectories.
The contrastive objective is
\begin{equation}
\mathcal L_{\mathrm{NCE}}
=
-
\log
\frac{
\exp(\operatorname{sim}(z_i,z_i^+)/\eta)
}{
\sum_j
\exp(\operatorname{sim}(z_i,z_j)/\eta)
},
\label{eq:idi_nce}
\end{equation}
where $\operatorname{sim}(\cdot,\cdot)$ denotes cosine similarity and $\eta$ is a temperature parameter \cite{rusak2024infonce}.
The resulting representation captures temporal compatibility over recent trajectory segments and enables support estimation through local neighborhood structure.

Let $\mathcal Z_{\mathcal D}=\{z_j\}$ denote the embeddings of all offline trajectory windows.
For an imagined post-transition window $\hat w_t$, we compute $\hat z_t=f_\phi(\hat w_t)$ and define its support score as
\begin{equation}
d_{\mathcal D}(\hat w_t)
=
\frac{1}{K}
\sum_{z_j\in\mathcal N_K(\hat z_t;\mathcal Z_{\mathcal D})}
\|\hat z_t-z_j\|_2,
\label{eq:idi_support}
\end{equation}
where $\mathcal N_K(\hat z_t;\mathcal Z_{\mathcal D})$ denotes the $K$ nearest offline embeddings.
Large values of $d_{\mathcal D}(\hat w_t)$ indicate that the candidate transition is incompatible with trajectory segments observed in the offline dataset.

We determine the support threshold using offline windows.
For each offline embedding $z_j$, its leave-one-out support score is
\begin{equation}
d_{\mathcal D}^{\mathrm{LOO}}(w_j)
=
\frac{1}{K}
\sum_{z_\ell\in
\mathcal N_K(z_j;\mathcal Z_{\mathcal D}\setminus\{z_j\})}
\|z_j-z_\ell\|_2.
\label{eq:idi_loo_support}
\end{equation}
The threshold is then defined as
\begin{equation}
c
=
\operatorname{Quantile}_q
\left(
\left\{
d_{\mathcal D}^{\mathrm{LOO}}(w_j)
\right\}_{w_j\in\mathcal D}
\right).
\label{eq:idi_threshold}
\end{equation}
Excluding the query embedding itself prevents artificially small offline support scores.

Each imagined rollout is initialized at an offline state with at least $k-1$ preceding transitions in the same trajectory.
These preceding offline transitions provide the initial history, after which they are progressively replaced by imagined transitions.
At rollout step $t$, IDI first samples
\begin{equation}
\hat a_t\sim\pi_t(\cdot|\hat s_t),
\qquad
\hat s_{t+1}
\sim
\hat P_\theta(\cdot|\hat s_t,\hat a_t),
\label{eq:idi_candidate_transition}
\end{equation}
and forms the post-transition window $\hat w_t$ containing the candidate transition $(\hat s_t,\hat a_t,\hat s_{t+1})$.
The transition is accepted if
\begin{equation}
d_{\mathcal D}(\hat w_t)\le c.
\label{eq:idi_acceptance}
\end{equation}
If \eqref{eq:idi_acceptance} holds, the transition is added to the synthetic dataset and imagination continues from $\hat s_{t+1}$.
Otherwise, the candidate transition is discarded and the rollout terminates.
Thus, an unsupported transition is rejected before it can enter the policy-training dataset.

We combine IDI with TRAC, which formulates finite-horizon offline RL as
\begin{equation}
\max_\pi
\;
\mathbb E_{\tau\sim P_\pi}[R(\tau)]
-
\lambda
D_{\mathrm{KL}}
\left(
P_\pi
\;\|\;
P_\mu
\right),
\label{eq:idi_trac_objective}
\end{equation}
where $P_\pi$ and $P_\mu$ are the trajectory distributions induced by the learned policy and the offline behavior policy, respectively.
TRAC uses the behavior policy as an action prior and yields a behavior-prior Bellman recursion and behavior-prior Boltzmann policy.
When the action prior is uniform, the formulation recovers finite-horizon maximum-entropy RL up to a policy-independent additive constant.
Appendix~\ref{app:trac} provides the complete formulation.

For each accepted transition, we compute $\hat r_t=r(\hat s_t,\hat a_t)$ and construct
\begin{equation}
\mathcal D_{\mathrm{IDI}}
=
\left\{
(\hat s_t,\hat a_t,\hat r_t,\hat s_{t+1})
:
d_{\mathcal D}(\hat w_t)\le c
\right\}.
\label{eq:idi_dataset}
\end{equation}
The actor and critics are trained on
$\mathcal D_{\mathrm{train}}=\mathcal D\cup\mathcal D_{\mathrm{IDI}}$, while the TRAC behavior prior remains trained solely on the original offline dataset $\mathcal D$.
IDI truncation is not treated as an environment terminal condition: the final accepted transition continues to bootstrap unless it reaches a true finite-horizon terminal state.

In our experiments, IDI retains approximately $55$--$70\%$ of generated transitions, yielding a synthetic dataset comparable in size to the original limited-data dataset.
By rejecting unsupported transitions before policy optimization, IDI improves the reliability of imagined data while TRAC constrains policy improvement toward behavior-supported actions.

\begin{figure}[t]

\centering
\includegraphics[width=\linewidth]{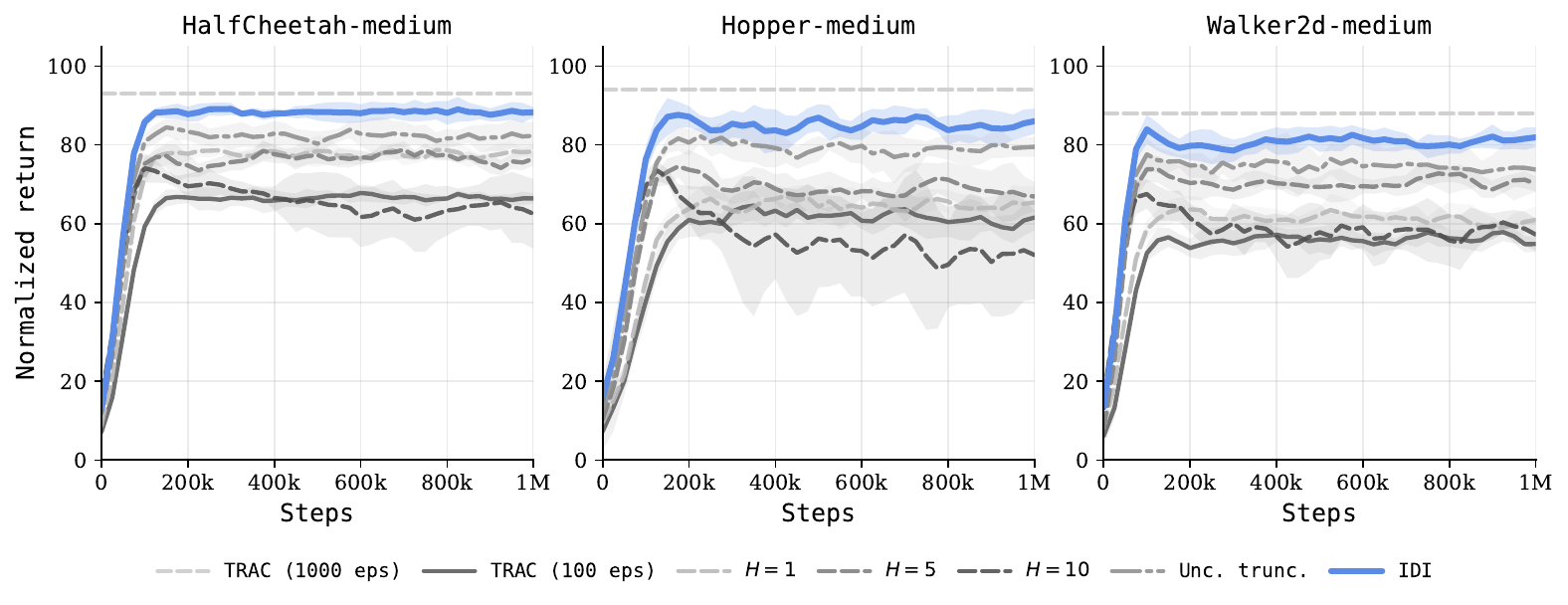}

\caption{
Normalized return learning curves on MuJoCo locomotion benchmarks under the limited-data setting.
TRAC is trained using only $100$ episodes, while the dashed line denotes TRAC trained on the full dataset ($1000$ episodes).
Fixed-horizon and uncertainty-based truncation improve over the limited-data baseline, but their effectiveness varies across environments.
IDI consistently achieves the strongest performance through history-aware rollout control.
}
\label{fig:learning_curves}
\end{figure}

%%%%%%%%%%%%%%%%%%%%%%%%%%%%%%%%%%%%%%%%%%%%%%%%%%%%%%%%%%%%%%%%
%% Section: Experiments
%%%%%%%%%%%%%%%%%%%%%%%%%%%%%%%%%%%%%%%%%%%%%%%%%%%%%%%%%%%%%%%%

\section{Experiments}
\label{sec:experiments}

We evaluate whether IDI improves MBORL in a limited-data regime and whether trajectory support predicts rollout failure.
For each MuJoCo dataset, we subsample $100$ episodes from the original $1000$ and train both the dynamics model and policy only using the subsampled data.
Experiments are conducted on \texttt{halfcheetah-medium-v0}, \texttt{hopper-medium-v0}, and \texttt{walker2d-medium-v0} \cite{fu2020d4rl, minari}.
All methods use the TRAC backbone and learned dynamics model.
We compare: (i) full-data TRAC ($1000$ episodes), (ii) limited-data TRAC ($100$ episodes), (iii) fixed-horizon rollouts, (iv) uncertainty-based truncation, and (v) IDI.
For uncertainty truncation, thresholds are selected from the same quantile grid used by IDI for fair comparisons.

\paragraph{Limited-data performance.}

Figure~\ref{fig:learning_curves} shows normalized return learning curves.
Limited-data TRAC suffers a substantial performance drop relative to full-data TRAC.
Model-generated trajectories partially recover this gap, but performance depends strongly on rollout control.
Fixed horizons provide inconsistent gains across environments, while uncertainty-based truncation is limited by noisy local uncertainty estimates.
IDI consistently achieves the strongest performance.
For example, on \texttt{halfcheetah-medium-v0}, IDI improves normalized return from approximately $67$ to $88$, approaching the full-data performance of approximately $93$.
Unlike fixed-horizon rollouts, IDI adapts rollout length to rollout reliability.
The average accepted rollout horizon is $6.2$, $4.8$, and $3.7$ steps on HalfCheetah, Hopper, and Walker2d, respectively, indicating that no single rollout horizon is optimal across environments.
These results suggest that synthetic data quality, rather than quantity, is the primary factor determining MBORL performance.

\paragraph{Rollout failure detection.}

\begin{figure}[t]

\centering

\begin{minipage}[t]{0.64\linewidth}
    \centering
    \includegraphics[width=\linewidth]{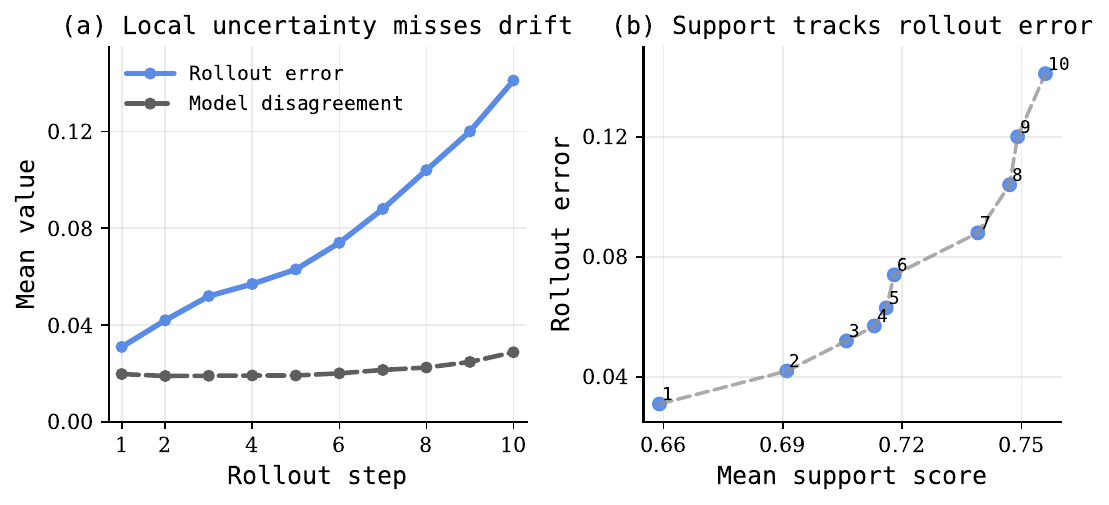}
    \caption{
    Rollout failure diagnostics on \texttt{halfcheetah}.
    (a) Rollout error and ensemble disagreement versus rollout horizon.
    (b) Trajectory support score versus rollout error.
    Support exhibits a strong monotonic relationship with rollout error.
    }
    \label{fig:support_diagnostics}
\end{minipage}
\hfill
\begin{minipage}[t]{0.34\linewidth}
    \vspace{-11.3em}
    \centering
    \small
    \begin{tabular}{lcc}
    \toprule
    Score & Corr. & AUROC \\
    \midrule
    Disagreement & 0.31 & 0.63 \\
    One-step error & 0.58 & 0.72 \\
    Window KNN & 0.79 & 0.84 \\
    IDI support & \textbf{0.94} & \textbf{0.91} \\
    \bottomrule
    \end{tabular}
    \captionof{table}{
    Quantitative rollout failure detection.
    Correlation is computed with open-loop rollout error and AUROC detects high-error rollouts.
    Both measures imply that IDI framework successfully detects trajectory drift and only accepts realistic rollouts.
    }
    \label{tab:failure_detection}
\end{minipage}

\end{figure}

We next evaluate whether trajectory support successfully predicts rollout failure.
Using the learned dynamics model on \texttt{halfcheetah-medium-v0}, we generate open-loop rollouts initialized from held-out offline states.
Rollout error is measured as $e_t=\mathbb E[\|\hat s_t-s_t\|_1]$ and ensemble disagreement as $u_t=\mathbb E[\mathrm{Std}(\hat s_t^{(1)},\dots,\hat s_t^{(N)})]$.
Trajectory support is measured by $d_{\mathcal D}(\hat w_t)$ and averaged across rollouts.
In Figure~\ref{fig:support_diagnostics}(b), each point corresponds to one rollout step.
Figure~\ref{fig:support_diagnostics}(a) shows that rollout error increases from approximately $0.03$ to $0.14$ over the first ten rollout steps, while ensemble disagreement changes only modestly.
Figure~\ref{fig:support_diagnostics}(b) shows that trajectory support increases nearly monotonically with rollout error, from approximately $0.66$ to $0.76$.
Notably, support continues to increase even when ensemble disagreement remains nearly constant, indicating that accumulated rollout drift is not fully reflected by local uncertainty estimates.
Together with the performance gains in Figure~\ref{fig:learning_curves}, these results suggest that rollout reliability is better characterized by trajectory support than transition-level uncertainty.

%%%%%%%%%%%%%%%%%%%%%%%%%%%%%%%%%%%%%%%%%%%%%%%%%%%%%%%%%%%%%%%%
%% Section: Conclusion
%%%%%%%%%%%%%%%%%%%%%%%%%%%%%%%%%%%%%%%%%%%%%%%%%%%%%%%%%%%%%%%%

\section{Conclusion}

We introduced IDI, a history-aware rollout control framework for model-based offline reinforcement learning. IDI represents recent trajectory segments in a learned latent space, evaluates each candidate transition using its post-transition window, and terminates imagination before unsupported transitions enter the policy-training dataset. This provides an adaptive alternative to fixed rollout horizons and local ensemble-based uncertainty estimates. Combined with TRAC, the two components address complementary sources of distribution shift: IDI controls the reliability of model-generated data, while TRAC regularizes policy improvement toward behavior-supported actions.
Experiments in limited-data MuJoCo settings demonstrate that IDI consistently improves performance over fixed-horizon and uncertainty-based rollout control. The learned support score also exhibits a substantially stronger relationship with accumulated rollout error than instantaneous ensemble disagreement, suggesting that recent rollout history provides information not captured by local uncertainty alone. These findings support a broader view of reliable imagination as a sequential data-selection problem: model-generated transitions should be evaluated not only individually, but also by their compatibility with the trajectory prefix that produced them. Future work may extend this principle to higher-dimensional observations, jointly learned world-model representations, and adaptive support thresholds that account for task relevance and model improvement.

%%%%%%%%%%%%%%%%%%%%%%%%%%%%%%%%%%%%%%%%%%%%%%%%%%%%%%%%%%%%%%%%
%% Appendices
%%%%%%%%%%%%%%%%%%%%%%%%%%%%%%%%%%%%%%%%%%%%%%%%%%%%%%%%%%%%%%%%
\newpage
\appendix

\section{Trajectory-Regularized Actor-Critic (TRAC)}
\label{app:trac}

TRAC is the offline RL backbone used throughout this paper. It formulates offline RL as finite-horizon KL control by regularizing the policy-induced trajectory distribution toward the behavior trajectory distribution~\cite{kim2026trajectory, kim2026finite}.

Consider a finite-horizon MDP $\mathcal M=(\mathcal S,\mathcal A,p,r,r_H,H,\rho_0)$ with trajectory $\tau=(s_0,a_0,\ldots,s_{H-1},a_{H-1},s_H)$ and return
$R(\tau)=\sum_{t=0}^{H-1}r_t+r_H$, where $r_t:=r(s_t,a_t)$ and $r_H:=r_H(s_H)$.
A nonstationary policy $\pi=(\pi_0,\ldots,\pi_{H-1})$ induces the trajectory distribution
\begin{equation}
P_\pi(\tau)
=
\rho_0(s_0)
\prod_{t=0}^{H-1}
\pi_t(a_t|s_t)
p(s_{t+1}|s_t,a_t).
\end{equation}
Let $\mu=(\mu_0,\ldots,\mu_{H-1})$ denote the behavior policy underlying the offline dataset and let $P_\mu$ denote its trajectory distribution. TRAC solves
\begin{equation}
\max_{\pi}
\;
\mathbb E_{\tau\sim P_\pi}[R(\tau)]
-
\lambda
D_{\mathrm{KL}}
\!\left(
P_\pi
\;\|\;
P_\mu
\right),
\label{eq:trac_objective}
\end{equation}
where $\lambda>0$ controls the strength of trajectory regularization. We assume $\pi_t(\cdot|s)\ll\mu_t(\cdot|s)$ on all states reachable under $\pi$, ensuring that the trajectory KL is finite.

Because $P_\pi$ and $P_\mu$ share the same initial-state distribution and transition dynamics, their likelihood ratio depends only on their action distributions. Consequently,
\begin{equation}
D_{\mathrm{KL}}
\!\left(
P_\pi
\;\|\;
P_\mu
\right)
=
\mathbb E_{\tau\sim P_\pi}
\left[
\sum_{t=0}^{H-1}
\log
\frac{\pi_t(a_t|s_t)}
{\mu_t(a_t|s_t)}
\right].
\label{eq:trac_kl_decomposition}
\end{equation}
Substituting \eqref{eq:trac_kl_decomposition} into \eqref{eq:trac_objective} gives
\begin{equation}
\max_{\pi}
\;
\mathbb E_{\tau\sim P_\pi}
\left[
\sum_{t=0}^{H-1}
\left(
r_t
-
\lambda
\log
\frac{\pi_t(a_t|s_t)}
{\mu_t(a_t|s_t)}
\right)
+
r_H
\right].
\label{eq:trac_decomposed_objective}
\end{equation}
Unlike a discounted infinite-horizon objective, both the return and trajectory regularization in \eqref{eq:trac_decomposed_objective} are undiscounted over the finite horizon.

The objective admits a behavior-prior Bellman recursion. Define the terminal value as
\begin{equation}
V_H(s)=r_H(s).
\end{equation}
For $t=H-1,\ldots,0$, define
\begin{equation}
Q_t(s,a)
=
r(s,a)
+
\mathbb E_{s'\sim p(\cdot|s,a)}
\left[
V_{t+1}(s')
\right].
\label{eq:trac_q_recursion}
\end{equation}
The regularized state value satisfies the state-wise variational problem
\begin{equation}
V_t(s)
=
\sup_{\pi_t(\cdot|s)\ll\mu_t(\cdot|s)}
\mathbb E_{a\sim\pi_t(\cdot|s)}
\left[
Q_t(s,a)
-
\lambda
\log
\frac{\pi_t(a|s)}
{\mu_t(a|s)}
\right].
\label{eq:trac_variational_value}
\end{equation}
Applying the Gibbs variational principle yields
\begin{equation}
V_t(s)
=
\lambda
\log
\mathbb E_{a\sim\mu_t(\cdot|s)}
\left[
\exp
\left(
\frac{Q_t(s,a)}{\lambda}
\right)
\right].
\label{eq:trac_behavior_prior_value}
\end{equation}
The optimizer is the behavior-prior Boltzmann policy
\begin{equation}
\pi_t^*(a|s)
=
\mu_t(a|s)
\exp
\left(
\frac{Q_t(s,a)-V_t(s)}{\lambda}
\right)
\propto
\mu_t(a|s)
\exp
\left(
\frac{Q_t(s,a)}{\lambda}
\right).
\label{eq:trac_optimal_policy}
\end{equation}
Thus, policy improvement reweights behavior-supported actions according to their values, while actions outside the behavior support remain excluded.

In offline learning, the behavior policy is unknown and is approximated by a behavior prior $\hat\mu_\phi$. The prior is learned only from the original offline dataset:
\begin{equation}
\phi^*
=
\arg\max_{\phi}
\;
\mathbb E_{(t,s_t,a_t)\sim\mathcal D}
\left[
\log
\hat\mu_{\phi,t}(a_t|s_t)
\right].
\label{eq:trac_behavior_prior}
\end{equation}
When a stationary prior is used in practice, we set $\hat\mu_{\phi,t}=\hat\mu_\phi$ for all $t$.

The practical actor update is the reverse-KL projection onto the behavior-prior Boltzmann policy. Equivalently, the actor minimizes
\begin{equation}
\mathcal L_\pi
=
\mathbb E_{\substack{(t,s_t)\sim\mathcal D_{\mathrm{train}}\\
a_t\sim\pi_t(\cdot|s_t)}}
\left[
\lambda
\log
\frac{\pi_t(a_t|s_t)}
{\hat\mu_{\phi,t}(a_t|s_t)}
-
Q_t(s_t,a_t)
\right].
\label{eq:trac_actor_loss}
\end{equation}
The critic is trained using the undiscounted finite-horizon target
\begin{equation}
y_t
=
\begin{cases}
r_t+\bar V_{t+1}(s_{t+1}), & t<H-1,\\
r_{H-1}+r_H(s_H), & t=H-1,
\end{cases}
\label{eq:trac_critic_target}
\end{equation}
where the target value is computed using the learned behavior prior:
\begin{equation}
\bar V_{t+1}(s')
=
\lambda
\log
\mathbb E_{a'\sim\hat\mu_{\phi,t+1}(\cdot|s')}
\left[
\exp
\left(
\frac{\bar Q_{t+1}(s',a')}{\lambda}
\right)
\right].
\label{eq:trac_practical_value}
\end{equation}
For continuous actions, the expectation in \eqref{eq:trac_practical_value} is approximated using samples $a'_1,\ldots,a'_M\sim\hat\mu_{\phi,t+1}(\cdot|s')$:
\begin{equation}
\bar V_{t+1}(s')
\approx
\lambda
\log
\left[
\frac{1}{M}
\sum_{m=1}^{M}
\exp
\left(
\frac{\min_{j\in\{1,2\}}\bar Q_{t+1}^{(j)}(s',a'_m)}
{\lambda}
\right)
\right].
\label{eq:trac_mc_value}
\end{equation}
The two target critics mitigate value overestimation and are updated using Polyak averaging. In implementation, the log-sum-exp in \eqref{eq:trac_mc_value} is evaluated with max subtraction for numerical stability. The stage indices describe the exact finite-horizon recursion and do not require separate networks at every stage; actor and critic parameters may be shared across stages.

TRAC recovers finite-horizon maximum-entropy RL when the behavior prior is replaced by a uniform action prior, up to a policy-independent additive constant. TRAC therefore generalizes entropy regularization by replacing its uniform action reference with a data-dependent behavior prior.

In our method, IDI is applied before policy optimization. Accepted imagined transitions form
$\mathcal D_{\mathrm{train}}=\mathcal D\cup\mathcal D_{\mathrm{IDI}}$, and the TRAC actor and critics are trained using $\mathcal D_{\mathrm{train}}$. The behavior prior $\hat\mu_\phi$, however, remains trained solely on the original offline dataset $\mathcal D$, preserving the intended trajectory reference. IDI rollout truncation is not treated as an environment terminal condition; the final accepted transition continues to bootstrap unless it coincides with a true finite-horizon terminal state. Consequently, IDI controls the reliability of imagined data, while TRAC regularizes policy optimization toward behavior-supported actions.

%%%%%%%%%%%%%%%%%%%%%%%%%%%%%%%%%%%%%%%%%%%%%%%%%%%%%%%%%%%%%%%%
%% Appendix B: Implementation Details
%%%%%%%%%%%%%%%%%%%%%%%%%%%%%%%%%%%%%%%%%%%%%%%%%%%%%%%%%%%%%%%%

\section{Implementation Details}
\label{app:implementation}

Experiments are conducted on \texttt{halfcheetah-medium-v0}, \texttt{hopper-medium-v0}, and \texttt{walker2d-medium-v0}.
Following the limited-data setting in Section~\ref{sec:experiments}, all methods are trained using $100$ trajectories randomly subsampled from the original dataset.
The same subsampled dataset is used to train the dynamics model, behavior prior, and trajectory encoder for all rollout-control methods.

\paragraph{Transition dynamics modeling.}
We use a probabilistic ensemble of $N=7$ Gaussian neural networks \cite{chua2018deep, cai2025learning, gupta2025estimation, kim2025roverfly}.
Each network is a three-layer MLP with hidden dimension $256$ and SiLU activations.
The models are trained by maximum likelihood using transitions from the original offline dataset.
During rollout generation, one ensemble member is sampled uniformly at each step to generate the candidate next state.

\paragraph{Trajectory representation.}
The trajectory encoder is a two-layer MLP with hidden dimension $256$ and latent dimension $64$.
Each state and action dimension is standardized using statistics computed from the original offline dataset, and the same transformation is applied to imagined transitions.
We use a window size of $k=8$ transitions, so each encoder input contains nine states and eight actions:
\begin{equation}
w_t
=
(s_{t-7},a_{t-7},\ldots,s_t,a_t,s_{t+1}).
\end{equation}
The normalized window is flattened and passed through the encoder.
The encoder is trained on offline trajectory windows using InfoNCE with temperature $\eta=0.1$.
Positive pairs consist of nearby windows from the same trajectory, while negative windows are sampled from different trajectories.
After training, the encoder is frozen and used for all support computations and imagined rollouts.

\paragraph{IDI.}
IDI uses $K=10$ nearest neighbors and a support threshold corresponding to the $95$th percentile of offline leave-one-out support scores.
When computing an offline window's support score, the query embedding itself is excluded from the KNN reference set.

Rollouts are initialized from offline states that have at least $k-1$ preceding transitions in the same trajectory.
These preceding offline transitions provide the initial history.
Consequently, the first imagined window contains $k-1$ offline transitions and one candidate imagined transition.
As imagination proceeds, accepted imagined transitions progressively replace the offline transitions in the window.

At each rollout step, IDI first samples a candidate action and next state:
\begin{equation}
\hat a_t\sim\pi_t(\cdot|\hat s_t),
\qquad
\hat s_{t+1}
\sim
\hat P_\theta(\cdot|\hat s_t,\hat a_t).
\end{equation}
It then forms the post-transition window $\hat w_t$, which includes the candidate transition $(\hat s_t,\hat a_t,\hat s_{t+1})$, and computes $d_{\mathcal D}(\hat w_t)$.
If $d_{\mathcal D}(\hat w_t)\le c$, the transition is accepted, added to $\mathcal D_{\mathrm{IDI}}$, and used as part of the history at the next rollout step.
If $d_{\mathcal D}(\hat w_t)>c$, the candidate transition is discarded and the rollout terminates.
Thus, a transition that crosses the support threshold never enters the policy-training dataset.

Rollouts are capped at a maximum horizon of $10$ steps.
IDI truncation and the maximum rollout cap are not treated as environment terminal conditions.
The final accepted transition continues to bootstrap unless it coincides with the true finite-horizon terminal state.
Across environments, IDI accepts approximately $55$--$70\%$ of generated transitions, corresponding to average accepted rollout horizons of $6.2$, $4.8$, and $3.7$ steps on HalfCheetah, Hopper, and Walker2d, respectively.

\paragraph{Policy optimization.}
The TRAC actor, behavior prior, and double critics use two-layer MLPs with hidden dimension $256$.
The behavior prior is trained by behavior cloning using only the original offline dataset $\mathcal D$ and remains fixed during policy optimization.
The actor and critics are trained on
$\mathcal D_{\mathrm{train}}=\mathcal D\cup\mathcal D_{\mathrm{IDI}}$.
Actor and critic parameters are shared across finite-horizon stages rather than using a separate network at each stage.

We use Adam with learning rate $3\times10^{-4}$, batch size $256$, and target-network update coefficient $\tau=0.005$.
Consistent with the finite-horizon TRAC formulation in Appendix~\ref{app:trac}, critic targets are undiscounted and no discount factor is applied.
Bootstrapping is removed only at the true finite-horizon terminal state.
The TRAC regularization coefficient $\lambda$ is learned automatically; no separate entropy coefficient is used.
All reported results are averaged over five random seeds.

%%%%%%%%%%%%%%%%%%%%%%%%%%%%%%%%%%%%%%%%%%%%%%%%%%%%%%%%%%%%%%%%
%% Appendix C: Additional Performance Results
%%%%%%%%%%%%%%%%%%%%%%%%%%%%%%%%%%%%%%%%%%%%%%%%%%%%%%%%%%%%%%%%

\section{Additional Performance Results}
\label{app:performance}

This section provides complete numerical results corresponding to Figure~\ref{fig:learning_curves}. All values report normalized return averaged over five random seeds with one standard deviation.

\paragraph{Final Performance.}

Table~\ref{tab:full_performance} reports the final normalized return achieved by each method under the limited-data setting. Full-data TRAC serves as an upper-reference performance level, while TRAC trained on only $100$ trajectories serves as the limited-data baseline.

\begin{table}[t]
\centering
\caption{
Final normalized return (mean $\pm$ std) under the limited-data setting.
}
\label{tab:full_performance}
\small
\begin{tabular}{lccc}
\toprule
Method &
HalfCheetah &
Hopper &
Walker2d \\
\midrule
TRAC (1000 traj.) &
$\mathbf{93.2 \pm 1.1}$ &
$\mathbf{91.5 \pm 2.3}$ &
$\mathbf{88.7 \pm 1.9}$ \\

TRAC (100 traj.) &
$67.4 \pm 3.6$ &
$61.2 \pm 5.4$ &
$58.7 \pm 4.2$ \\

Fixed $H=1$ &
$74.5 \pm 2.8$ &
$67.9 \pm 4.7$ &
$65.1 \pm 3.6$ \\

Fixed $H=3$ &
$81.7 \pm 2.5$ &
$74.3 \pm 3.8$ &
$71.8 \pm 3.1$ \\

Fixed $H=5$ &
$84.6 \pm 2.1$ &
$79.8 \pm 3.1$ &
$75.2 \pm 2.7$ \\

Fixed $H=10$ &
$78.9 \pm 8.7$ &
$72.1 \pm 10.3$ &
$69.4 \pm 9.1$ \\

Uncertainty truncation &
$86.4 \pm 2.6$ &
$81.7 \pm 3.3$ &
$77.8 \pm 3.0$ \\

IDI &
$\mathbf{88.1 \pm 1.8}$ &
$\mathbf{85.4 \pm 2.6}$ &
$\mathbf{81.9 \pm 2.4}$ \\
\bottomrule
\end{tabular}
\end{table}

IDI consistently achieves the strongest performance among rollout-control methods across all environments. Fixed-horizon rollouts improve upon the limited-data baseline but exhibit substantial sensitivity to the rollout horizon. In particular, $H=10$ often introduces unstable training due to compounding model error, resulting in significantly larger variance. Uncertainty-based truncation reduces some rollout failures, but remains less effective than trajectory-level support estimation.

\paragraph{Data Efficiency Recovery.}

To quantify how much of the lost performance is recovered through synthetic data generation, we define the recovery ratio

\begin{equation}
R
=
\frac{
J_{\mathrm{method}}
-
J_{\mathrm{TRAC100}}
}{
J_{\mathrm{TRAC1000}}
-
J_{\mathrm{TRAC100}}
}.
\end{equation}

Table~\ref{tab:recovery_ratio} reports recovery ratios for the strongest rollout-control methods.

\begin{table}[t]
\centering
\caption{
Fraction of lost performance recovered relative to full-data TRAC.
}
\label{tab:recovery_ratio}
\small
\begin{tabular}{lccc}
\toprule
Method &
HalfCheetah &
Hopper &
Walker2d \\
\midrule
Fixed $H=5$ &
0.67 &
0.61 &
0.55 \\

Uncertainty truncation &
0.74 &
0.68 &
0.64 \\

IDI &
\textbf{0.80} &
\textbf{0.80} &
\textbf{0.77} \\
\bottomrule
\end{tabular}
\end{table}

These results suggest that the benefit of synthetic data depends primarily on rollout quality rather than rollout quantity. IDI recovers a substantially larger fraction of the performance lost due to limited offline coverage by preventing unreliable long-horizon imagination.

%%%%%%%%%%%%%%%%%%%%%%%%%%%%%%%%%%%%%%%%%%%%%%%%%%%%%%%%%%%%%%%%
%%%%%%%%%%%%%%%%%%%%%%%%%%%%%%%%%%%%%%%%%%%%%%%%%%%%%%%%%%%%%%%%

% \label{sec:ack}
% Use unnumbered third level headings for the acknowledgments. All acknowledgments, including those to funding agencies, go at the end of the paper. Only add this information once your submission is accepted and deanonymized. The acknowledgments do not count towards the 8--12 page limit.

%%%%%%%%%%%%%%%%%%%%%%%%%%%%%%%%%%%%%%%%%%%%%%%%%%%%%%%%%%%%%%%%
%% NOTE: THIS MARKS THE END OF THE "MAIN TEXT"
%%%%%%%%%%%%%%%%%%%%%%%%%%%%%%%%%%%%%%%%%%%%%%%%%%%%%%%%%%%%%%%%

%%%%%%%%%%%%%%%%%%%%%%%%%%%%%%%%%%%%%%%%%%%%%%%%%%%%%%%%%%%%%%%%
%% Bibliography
%%%%%%%%%%%%%%%%%%%%%%%%%%%%%%%%%%%%%%%%%%%%%%%%%%%%%%%%%%%%%%%%
\newpage
\bibliography{main}
\bibliographystyle{rlj}

%%%%%%%%%%%%%%%%%%%%%%%%%%%%%%%%%%%%%%%%%%%%%%%%%%%%%%%%%%%%%%%%
% AUTHOR: If your paper has no supplementary materials, you may 
%         comment out the line below, which creates the title for
%         the supplementary materials.
%%%%%%%%%%%%%%%%%%%%%%%%%%%%%%%%%%%%%%%%%%%%%%%%%%%%%%%%%%%%%%%%
% \beginSupplementaryMaterials

% Content that appears after the references are not part of the ``main text,'' have no page limits, are not necessarily reviewed, and should not contain any claims or material central to the paper. 
% %
% If your paper includes supplementary materials, use the \begin{center}
%     {\tt {\textbackslash}beginSupplementaryMaterials} 
% \end{center}
% command as in this example, which produces the title and disclaimer above. 
% %
% If your paper does not include supplementary materials, this command can be removed or commented out.

\end{document}